\documentclass{article}
\usepackage{PRIMEarxiv}
\usepackage{cite}
\usepackage{amsmath,amssymb,amsfonts}
\usepackage{lipsum}
\usepackage{graphicx} 
  
\usepackage{textcomp}
\usepackage{color}
\usepackage{xcolor}
\usepackage{array}
\usepackage{booktabs}
\usepackage{tabularx}

\usepackage{algorithm}
\usepackage{algorithmic}

\def\BibTeX{{\rm B\kern-.05em{\sc i\kern-.025em b}\kern-.08em
    T\kern-.1667em\lower.7ex\hbox{E}\kern-.125emX}}

\usepackage{fancyhdr}

\title{RefPentester: A Knowledge-Informed Self-Reflective Penetration Testing Framework Based on Large Language Models
}

\author{
 Hanzheng Dai \\Concordia University \\ Montr\'eal, Canada\\
\texttt{hanzheng.dai@mail.concordia.ca} \\
   \And
 Yuanliang Li \\
  Concordia University \\ Montr\'eal, Canada\\
\texttt{l\_yuanli@live.concordia.ca} \\
  \And
  Jun Yan \\
  Concordia University \\ Montr\'eal, Canada\\
\texttt{jun.yan@concordia.ca}
  \And
  Zhibo Zhang \\
  Polytechnique Montr\'eal\\Montr\'eal, Canada\\
  \texttt{zhibo.zhang@polymtl.ca}
  \thanks{This work was supported in part by the Natural Sciences and Engineering Research Council of Canada (NSERC) Discovery Grant and the Concordia University Research Chair program.}
}

\begin{document}
\maketitle
\begin{abstract}
Automated penetration testing (AutoPT) powered by large language models (LLMs) has gained attention for its ability to automate ethical hacking processes and identify vulnerabilities in target systems by leveraging the inherent knowledge of LLMs.
However, existing LLM-based AutoPT frameworks often underperform compared to human experts in challenging tasks for several reasons: the imbalanced knowledge used in LLM training, short-sightedness in the planning process, and hallucinations during command generation. Moreover, the trial-and-error nature of the PT process is constrained by existing frameworks lacking mechanisms to learn from previous failures, restricting adaptive improvement of PT strategies.
To address these limitations, we propose a knowledge-informed, self-reflective PT framework powered by LLMs, called RefPentester. This AutoPT framework is designed to assist human operators in identifying the current stage of the PT process, selecting appropriate tactics and techniques for each stage, choosing suggested actions, providing step-by-step operational guidance, and reflecting on and learning from previous failed operations.
We also modeled the PT process as a seven-state Stage Machine to integrate the proposed framework effectively.
The evaluation shows that RefPentester can successfully reveal credentials on Hack The Box's Sau machine, outperforming the baseline GPT-4o model by 16.7\%. Across PT stages, RefPentester also demonstrates superior success rates on PT stage transitions.
\end{abstract}


\section{Introduction}
Protecting computer systems has become increasingly essential in the rapidly evolving digital environment. Penetration Testing (PT) is a commonly used approach to evaluate the security status of computer systems by launching a sequence of authorized cyberattacks to identify vulnerabilities that adversaries might exploit~\cite{baloch2017ethical}. 
Traditional PT is a complex, multi-stage process encompassing reconnaissance, vulnerability assessment, exploitation, and reporting \cite{pentest-standard}. Each stage often requires specialized expertise and considerable time investment by security professionals, resulting in prolonged system downtime and higher operational costs~\cite{denis2016penetration}. Automated penetration testing (AutoPT) aims to automate the PT process to enhance the efficiency of vulnerability identification in computer systems and minimize manual interventions. 
Several tools and frameworks have been developed to automate different stages of the PT process, such as  Metasploit~\cite{kennedy2011metasploit} and OpenVAS~\cite{rahalkar2019openvas}. However, the common use of AutoPT tools primarily automates only a part of the PT process, requiring oversight from PT experts and lacking the capability to make long-term decisions to fully automate the PT process~\cite{li2023hierarchical}.

Recently, PT applications based on machine learning (ML) techniques, such as reinforcement learning (RL) and large language models (LLMs), have been increasingly adopted\cite{mckinnel2019systematic}. 
RL is a machine learning paradigm where an agent learns to make optimal decisions in an environment by interacting with it, receiving rewards or penalties for its actions to maximize cumulative rewards over time~\cite{wiering2012reinforcement}. 
LLMs are powerful AI models based on deep neural networks, such as the Transformer architecture, and trained on vast amounts of text data to understand, generate, and utilize human language. These models can handle a wide range of language-related tasks, such as text generation, translation, and question answering~\cite{zhao2023survey}.
Previous studies~\cite{barron2024domain} have shown that LLMs' domain-specific knowledge can be enhanced by vectorial databases, which provide curated knowledge to LLMs via knowledge-informed prompts, mitigating hallucinations and improving output accuracy. For example, Patrick et al.~\cite{lewis2020retrieval} introduced retrieval-augmented generation (RAG), a technique that integrates external knowledge base retrieval with LLM generation capabilities. RAG retrieves relevant information based on user input and combines it with the query to generate a response through an LLM.

In addition, recent studies have investigated how LLMs can correct their low-quality generations by reflecting on their previous responses, inspired by how humans improve their answers through reflective practices.
Shinn et al.~\cite{shinn2024reflexion} introduced Reflexion, a novel
technique for LLMs that can enhance LLMs' decision-making ability through in-context RL, where responses generated by LLMs are reflected using verbal rewards. As a PT process typically involves multiple trial-and-error attempts to explore and exploit the target system to compromise it (capture credentials), the PT process can be formulated as a partially observable Markov decision process (POMDP) \cite{ghanem2019reinforcement} within the realm of RL, aiming to maximize long-term reward. Similarly, when we replace the RL agent with an LLM agent, Reflexion can bring the trial-and-error mechanism into LLM-based PT, potentially mitigating the complex decision-making challenges faced in LLM-based PT.

Many studies have investigated the application of RL and LLMs in advanced AutoPT tools or frameworks. 
Zhenguo et al.~\cite{hu2020automated} proposed an automated penetration testing framework based on the Deep Q-Learning Network (DQN). It uses the Shodan search engine to collect real-world server data for building realistic network topologies, multi-host multi-stage vulnerability analysis (MulVAL) to generate attack trees, and depth-first Search (DFS) to create a simplified matrix for DQN training.
Yuanliang et al.~\cite{li2024knowledge} proposed a knowledge-informed AutoPT framework based on RL with Reward Machine (DRLRM-PT). This framework integrates cybersecurity domain knowledge into RL to guide PT, thereby improving the learning efficiency of PT policies and revealing the potential of integrating ML and human knowledge in AutoPT. 
Deng et al. introduced PentestGPT~\cite{deng2024pentestgpt}, an innovative LLM-powered PT framework designed to leverage reasoning abilities and the intrinsic knowledge of LLMs to plan the PT process by creating a PT task tree, generating PT operation commands for the current task, and allowing users to execute the commands with human-in-the-loop involvement.
Jiacen et al.~\cite{xu2024autoattacker} proposed Autoattacker. It leverages LLMs to carry out various attack tasks across different endpoint configurations, comprising four main components: summarizer, planner, navigator, and experience manager. By carefully designing prompt templates for each component and using an LLM jailbreaking technique, Autoattacker can effectively obtain attack commands from LLMs.

Despite these advances, existing LLM-based AutoPT frameworks still have some limitations that may restrict their performance.
\begin{enumerate}
    \item For existing LLM-based AutoPT frameworks, short-sighted planning and hallucinations in task decomposition ~\cite{ji2023survey} and command generation limit the effectiveness of LLMs in managing the overall PT process.
    \item Context loss is often encountered when LLMs require multiple rounds of generations with large amounts of information as input~\cite{vaswani2017attention}. Also, context window limitations and lack of persistent memory impede LLMs.
    \item Existing LLM-based PT frameworks lack mechanisms to learn from previous experiences, limiting their capacity for long-term decision-making~\cite{amodei2016concrete} in PT. This also significantly hinders their adaptability in diverse target environments~\cite{sommer2010outside}.
    \item Imbalanced training data during the pre-training phase of LLMs can cause performance decline in specific domains, such as the PT domain~\cite{ju2024mitigating}.
\end{enumerate}

\begin{figure*}[htbp] 
\centering
\includegraphics[width=0.9\columnwidth]{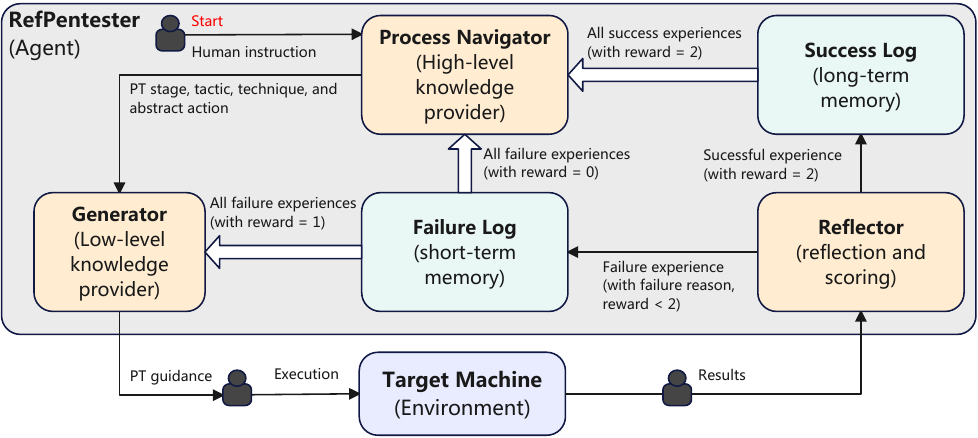}
\vspace{0mm}
\caption{The proposed RefPentester framework.}
\vspace{0mm}
\label{RefPentester_framework}
\end{figure*}

To address these challenges, we introduced RefPentester, a knowledge-informed, self-reflective AutoPT framework with human-in-the-loop based on LLMs. 
Our motivations are 4-fold: 1) To address LLMs' short-sighted planning and hallucinations, we designed a PT Stage Machine based on seven pre-defined PT stages. It can provide current PT stage by analyzing successful experiences;  2) To address LLMs' context loss, we designed a long-term memory mechanism to record all the successful experiences to make sure RefPentester will not forget the current stage of PT; 3) Through LLM-based reflection and the provision of in-context rewards and failure reasons, RefPentester can learn from past trial-and-error experiences via a short-term memory; 4) To tackle LLM performance degradation in PT domains caused by imbalanced training data, we utilize the RAG pipeline to offer pre-defined hierarchical PT knowledge.
In brief, the \textbf{contributions} of this work can be summarized as follows:
\begin{itemize}

\item We introduced RefPentester, a knowledge-informed AutoPT framework based on LLMs. It consists of multiple components. The Process Navigator utilizes a pre-defined PT Stage Machine to determine the current PT stage and acquire the high-level PT knowledge via an RAG pipeline. Subsequently, this high-level PT knowledge is fed to the Generator to generate detailed PT guidance. Finally, the Reflector evaluates and rewards both the PT guidance and high-level PT knowledge according to the execution results.

\item We gathered PT knowledge metadata from a wide range of public cybersecurity resources and constructed a high-level PT knowledge VDB capable of offering three-tier hierarchical PT knowledge under the current PT stage.

\item To illustrate the effectiveness of the proposed framework, we conducted a comparative case study on the Sau machine from the HTB platform using the base LLM model, GPT-4o. Results show our framework can find more credentials and boost the state-transition success rate in the PT process compared to the base model.
\end{itemize}




The rest of this paper is structured as follows: Section~\ref{section-Methodology} introduces the proposed RefPentester framework, including details of each component. Section~\ref{section-eval} describes the testing environment where comparative studies are performed to validate the performance of our framework. Conclusions and future work are presented in Section~\ref{section-con}.

\section{Methodology}\label{section-Methodology}

In this section, we provide an overview of the proposed RefPentester framework, followed by a detailed description of each component in it.

\begin{table}
\caption{List of Notations}
\vspace{0mm}
\begin{center}
\begin{tabular}{c c c c}
\hline
\specialrule{0em}{1pt}{1pt}
\textbf{Symbol} &\textbf{Description} &\textbf{Symbol} &\textbf{Description}\\
\specialrule{0em}{1pt}{1pt}
\hline
\specialrule{0em}{1pt}{1pt}

$t$ & Iteration & $q$ & Human Instruction\\ 
\specialrule{0em}{1pt}{1pt}

$\tau$ & PT Stage & $h$ & PT Event\\ 
\specialrule{0em}{1pt}{1pt}

$\mathcal{M}$ & PT Stage Machine & $\Pi$ & LLM Session\\ 
\specialrule{0em}{1pt}{1pt}

$c$ & Tactic & $\vec{c}$ & Vectorized Tactic\\ 
\specialrule{0em}{1pt}{1pt}

 $u$ & Technique & $\vec{u}$ & Vectorized Technique\\ 
\specialrule{0em}{1pt}{1pt}

$A$ & Potential Actions & $\vec{A}$ & Vectorized Potential Actions\\ 
\specialrule{0em}{1pt}{1pt}

$a$ & Abstract Action & $\vec{a}$ & Vectorized Abstract Action\\ 
\specialrule{0em}{1pt}{1pt}

$\mathcal{C}$ & Tactic Set & $\mathbb{V}_\mathcal{C}$ & Vectorized Tactic Set\\ 

$\mathcal{U}$ & Technique Set & $\mathbb{V}_\mathcal{U}$ & Vectorized Technique Set\\ 

$\mathcal{A}$ & Abstract Action Set & $\mathbb{V}_\mathcal{A}$ & Vectorized Abstract Action Set\\ 

$g$ & PT Guidance & $o$ & Results\\ 
\specialrule{0em}{1pt}{1pt}

$r$ & Reward & $\phi$ & Failure Reason\\ 
\specialrule{0em}{1pt}{1pt}

$Y$ & Success Log & $F$ & Failure Log\\ 
\specialrule{0em}{1pt}{1pt}

$y$ & Success Experience & $f$ & Failure Experience\\ 
\specialrule{0em}{1pt}{1pt}

$\Pi$ & LLM Session & $\mathcal{K}$ & RAG Pipeline\\ 
\specialrule{0em}{1pt}{1pt}

\specialrule{0em}{1pt}{1pt}

\specialrule{0em}{1pt}{1pt}

\hline
\end{tabular}
\vspace{1mm}
\label{notations}
\end{center}
\end{table}

\subsection{Framework Overview}
The proposed RefPentester framework consists of five components: \textbf{Process Navigator}, \textbf{Generator}, \textbf{Reflector}, \textbf{Success Log}, and \textbf{Failure Log}, as illustrated in Fig.~\ref{RefPentester_framework}, which plays as an assisted agent for the human operator to perform PT on a target machine. The notations of the framework are illustrated in Table~\ref{notations}.

The \textbf{Process Navigator} is a high-level PT knowledge provider powered by an RAG pipeline, which determines the high-level PT stage ($\tau$), tactic ($c$), technique ($u$) and abstract action ($a$) based on human input instructions ($q$). 
The \textbf{Generator} is a low-level knowledge provider, which can generate more detailed PT guidance ($g$) for the human operator to perform executions on the target machine.
The target machine will respond with results ($o$) to the human operator, who will input it to the Reflector.
The \textbf{Reflector} is used to evaluate and reflect on PT performance by assigning rewards ($r$) and identifying potential failure reasons for generated PT guidance and high-level PT knowledge based on $o$.
The \textbf{Success Log} ($Y$) is a list of successful experiences in temporal order. Each successful experience $y$ is a seven-element tuple: ($q, \tau, c, u, a, r, o$), which will be kept until the termination of the PT process.
The \textbf{Failure Log} ($F$) is a list of failure experiences in temporal order. Each failure experience $f$ is a nine-element tuple: ($q, \tau, c, u, a, r, g, o, \phi$), where $\phi$ represents failure reasons. $F$ will be reset if the executions corresponding to $a$ are executed successfully on the target machine.

Each component applies multiple LLM sessions (LLM session with a certain system prompt) to ensure that each session is responsible for a dedicated task during the PT process. 
All LLM sessions in each component are interconnected through a chaining structure, where one session's output serves as the input of the subsequent session. 

Initially, human operators provide $q$ to the Process Navigator, which will generate high-level PT knowledge ($\tau, c, u, a$) based on $q$, then provide the high-level knowledge to the Generator. The Generator will generate low-level PT knowledge $g$ to the human operator based on it, the human operator will follow the $g$ to execute executions on the target machine. The target machine will provide results $o$ to the Reflector. The Reflector will reflect and score high and low-level PT knowledge based on $o$. If the value of $r$ is two, it means the high and low-level knowledge are correct, recording the ($q, \tau, c, u, a, r, o$) as a successful experience to the $Y$. If the $r$ is less than two, that means that the execution based on PT guidance is failed, Reflector record the ($q, \tau, c, u, a, r, g, o, \phi$) to the $F$, and evaluate if the high-level knowledge ($c, u, a$) is correct. If it is correct, the value of the reward is one, we will go to the Generator to reflect on previous $g$; if it is not, the value of the reward is zero, and we will go to the Process Navigator to reflect on previous high-level PT knowledge. 

In the following sections, we introduce the PT knowledge preparation workflow for building the VDB and engineering processes for each component used in RefPentester.

\subsection{Knowledge Preparation}
\begin{figure}[htbp] 
\centering
\includegraphics[width=0.7\columnwidth]{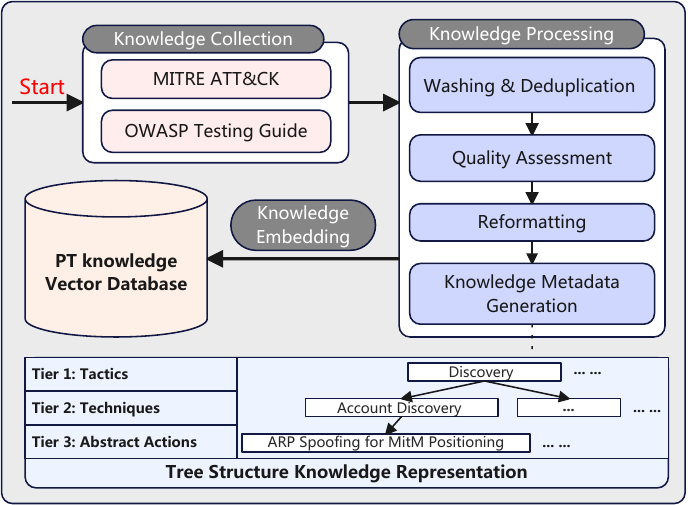}
\vspace{1mm}
\caption{PT knowledge preparation workflow for building a VDB.}
\vspace{1mm}
\label{knowledge preparing}
\end{figure}

The PT knowledge preparation workflow, depicted in Fig.~\ref{knowledge preparing}, creates a VDB that embeds PT knowledge through knowledge collection, processing, and embedding steps. For this purpose, we collect PT tactics, techniques, and actions from various public cybersecurity resources, including the MITRE ATT\&CK and the OWASP Testing Guide (OTG). 

MITRE ATT\&CK~\cite{mitreAttackMatrix} is a globally recognized knowledge base of adversary tactics, techniques, and procedures based on real-world observations. Tactics represent the high-level goals or objectives that an attacker aims to achieve during PT. Techniques are the methods employed by an adversary to achieve a specific tactical goal. Consequently, each tactic may leverage multiple techniques to achieve its intended outcome. However, for security reasons, the procedures defined in MITRE ATT\&CK are not specific, making them non-actionable. We cannot directly leverage it as more detailed PT action knowledge. 

OWASP Testing Guide (OTG) is a key resource for web server security testing~\cite{OWASPTestingGuideWebsite}. Using a black-box testing approach, OTG details security assessments across web servers and their deployment stacks with coverage of multiple aspects, like information gathering and configuration testing. We collect actionable and detailed PT knowledge from OTG and map it onto MITRE ATT\&CK techniques. For example, we collect OTG's Enumerate Applications on Webserver (OTG-INFO-004), then map it onto MITRE ATT\&CK's Gather Victim Host Information (T1592). This ensures that our PT knowledge is capable of supplying the LLM session with the action-level knowledge, thereby enabling the generation of the PT guidance for human operators to execute.

Based on these resources, our PT knowledge preparation workflow to build a VDB is illustrated in Fig.~\ref{knowledge preparing}.
Our PT knowledge representation is designed using a three-tiered tree structure. The tiers represent sets of tactics ($\mathcal{C}$), techniques ($\mathcal{U}$), and abstract actions ($\mathcal{A}$), respectively.
$\mathcal{C}$ and $\mathcal{U}$ are defined under the MITRE ATT\&CK Matrix for Enterprise. The definition of $\mathcal{A}$ is based on OTG. We adopted the following practices to collect PT knowledge:
\begin{enumerate}
    \item Provide precise guidance that offers meaningful and actionable insights.
    \item Refrain from providing overly specific details to ensure applicability across various contexts or scenarios.
    \item Adopt standard terminology within the MITRE ATT\&CK and the OTG to guarantee consistency and facilitate understanding throughout the security community.
\end{enumerate}

After the knowledge collection phase, we proceed to the knowledge processing phase. We first remove any redundant or inaccurate knowledge through washing and de-duplication. Then, a quality assessment is performed to evaluate the integrity and relevance of the knowledge.
After that, we reform the collected knowledge by adding the features of tactics and techniques to the knowledge of techniques and abstract actions, respectively. Since multiple child-level knowledge items correspond to each parent-level knowledge item, we combine the features of the parent-level knowledge item, such as tactic ID, with those of the child-level knowledge items. So we have the knowledge metadata with a three-tiered tree structure.
Finally, we embed the knowledge metadata as vectors and store them in one VDB. By doing so, we have the vectorized tactic, technique and abstract action set, which is notated as $\mathbb{V}_\mathcal{C}$, $\mathbb{V}_\mathcal{U}$ and $\mathbb{V}_\mathcal{A}$, respectively.
\subsection{Process Navigator}
The Process Navigator determines the high-level PT knowledge, which is informed by the PT Stage Machine ($\mathcal{M}$) and retrieved knowledge using the RAG pipeline ($\mathcal{K}$). First, it uses $\Pi_1$ and $\mathcal{M}$ to identify $\tau_t$. $t$ stands for iteration. Then, it applies the $\mathcal{K}$ to identify $c_{t}$, $u_{t}$, and potential abstract actions $A_{t}$. Finally, it uses another LLM session $\Pi_2$ to determine $a_t$.

\begin{figure}
\centering
\includegraphics[width=0.85\columnwidth]{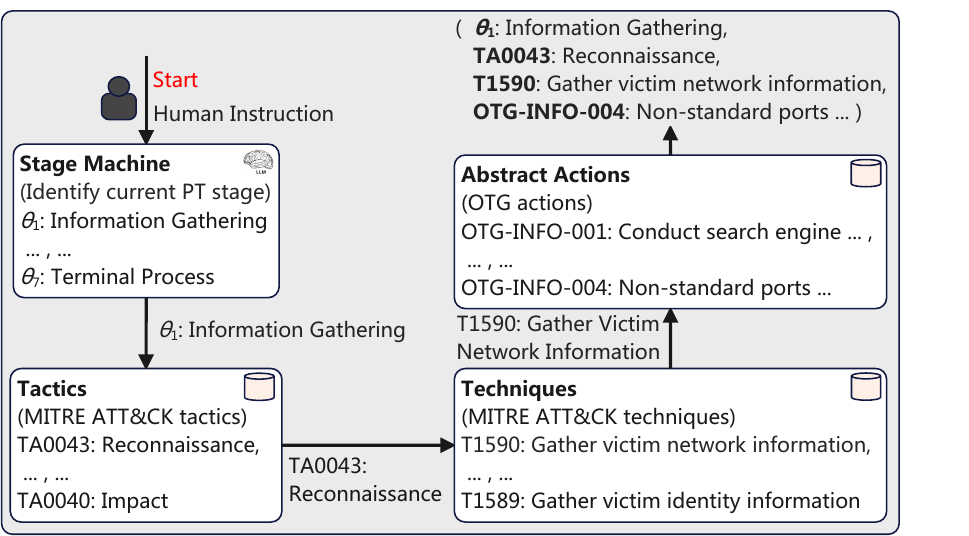}
\vspace{0mm}
\caption{An illustrative example of Process Navigator knowledge retrieval.}
\vspace{0mm}
\label{Knowledge Retrieve}
\end{figure}

The Process Navigator workflow is presented by Algorithm~\ref{Process Navigator Workflow}. The inputs include $q_t$, $Y_{t-1}$, and $F_{t-1}$. $q_t$ tells what the human operator intends to do during the current PT iteration. 
$Y_{t-1}$ records a list of successful experiences. These records are ranked in temporal sequence. By recording the Success Log of PT, our framework mitigates the risk of context loss inherent to LLMs, thereby preserving the continuity of the PT process. 
%
$F_{t-1}$ is a list of failure experiences that includes low-level and high-level PT knowledge corresponding to the unsuccessfully executed executions.
The failure experiences follow a temporal sequence. $F$ shall be reset if one PT action is successfully executed. By maintaining $F$ associated with the current failure execution, our framework can effectively leverage previous failure experiences to facilitate learning and improvement. 

\begin{algorithm}[htbp]
\caption{Process Navigator Workflow}
\label{Process Navigator Workflow}
\begin{algorithmic}[1]
\setlength{\baselineskip}{1.3\baselineskip}
\REQUIRE $q_t$, $Y_{t-1}$, $F_{t-1}$ \\
\textbf{Note:} Either $(Y_{t-1})$ or $(F_{t-1})$ should be empty.
\ENSURE $\tau_t$, $c_t$, $u_t$, $a_t$

\IF{$Y_{t-1}$ is not empty}
    \STATE $P_{t-1} \gets Y_{t-1}$
\ELSE
    \STATE $P_{t-1} \gets F_{t-1}$
\ENDIF

\STATE $h_t \gets \Pi_1(P_{t-1})$
\STATE $\tau_t \gets \mathcal{M}(h_t)$
\STATE $c_t, u_t, A_t \gets \mathcal{K}(q_t, \tau_t)$
\STATE $a_t \gets \begin{cases}\Pi_2(q_t\oplus P_{t-1}\oplus A_t) \quad & \text{if } k=0 \\\Pi_2(q_t\oplus P_{t-1}) \quad & \text{if } k=1\end{cases}$
\RETURN $\tau_t,c_t, u_t, a_t$
\end{algorithmic}
\end{algorithm}

Firstly, the Process Navigator applied $\Pi_1$ to identify the $h_t$. PT Stage Machine ($\mathcal{M}$) will determine the current $\tau_t$ by $h_t$. $\mathcal{M}$ is a state machine and will be introduced in the following subsection.
In the second step, the RAG pipeline ($\mathcal{K}$) will be applied to retrieve  $c_t$, $u_t$, and $A_t$ based on $q_t$ and the determined $\tau_t$. The detailed RAG pipeline will be introduced in the next part of the section.

Finally, $\Pi_2$ will be used to determine whether to pick an abstract action from $A_t$ or its inherent knowledge. Since accumulative reward is included in the logs ($Y_{t-1}$ and $F_{t-1}$), $\Pi_2$ will be leveraged to maximize the accumulative reward.
$k$ is an indicator that indicates $\Pi_2$ will choose an action from $A_t$ or generate a new action. If the $k$ is 0, that means that $\Pi_2$ will choose an action from $A_t$; if the $k$ is 1, that means that $\Pi_2$ think the actions inside $A_t$ is not suitable for current $q_t$ and Log ($Y_{t-1}$ or $F_{t-1}$), so $\Pi_2$ will leverage its inherent knowledge to generate a new abstract action. 
The workflow finalizes the operation by returning the $\tau_t$, $c_t$, $u_t$ and $a_t$. Fig.~\ref{Knowledge Retrieve} is an illustrative example of knowledge retrieval in Process Navigator. Initially, the PT Stage Machine will be used to identify the PT stage based on the PT event, which is determined by $\Pi_1$. In this case, the PT stage is $\theta_1$: information gathering. $\theta_1$ will be applied to retrieve tactic knowledge (TA0043: Reconnaissance) from VDB. After that, the Process Navigator will use the tactic knowledge and technique knowledge to retrieve the technique (T1590: Gather Victim Network Information) and abstract actions (OTG-INFO-001, etc), respectively.

\subsubsection{PT Stage Machine}
\begin{table}
\caption{PT Stages and events}
\vspace{0mm}
\begin{center}
\begin{tabular}{c c c c}
\hline
\specialrule{0em}{1pt}{1pt}
\textbf{Stage} &\textbf{Description} & \textbf{Event} & \textbf{Description}\\
\specialrule{0em}{1pt}{1pt}
\hline
\specialrule{0em}{1pt}{1pt}

$\theta_1$ & Information Gathering & $\alpha$ & Gathered Information\\ 
\specialrule{0em}{1pt}{1pt}

$\theta_2$ & Vulnerability Identification & $\beta$ & Identified Vulnerability\\ 
\specialrule{0em}{1pt}{1pt}

$\theta_3$ & Exploitation & $\gamma$ & Exploited\\ 
\specialrule{0em}{1pt}{1pt}

$\theta_4$ & Post-Exploitation & $\delta$ & Post-Exploited\\ 
\specialrule{0em}{1pt}{1pt}

$\theta_5$ & Capture the Flag & $\epsilon$ & Flag Captured\\ 
\specialrule{0em}{1pt}{1pt}

$\theta_6$ & Documentation & $\zeta$ & Documented\\ 
\specialrule{0em}{1pt}{1pt}

$\theta_7$ & Terminal Process &  & \\ 
\specialrule{0em}{1pt}{1pt}

\specialrule{0em}{1pt}{1pt}

\hline
\end{tabular}
\vspace{0mm}
\label{PT stages}
\end{center}
\end{table}

\begin{figure}
\centering
\includegraphics[width=0.4\columnwidth]{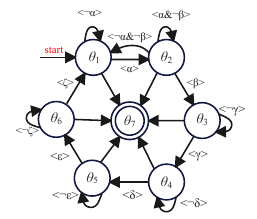}
\vspace{0mm}
\caption{The PT Stage Machine.}
\vspace{0mm}
\label{pt_stage}
\end{figure}

The PT Stage Machine is a state machine, as shown in Fig.~\ref{pt_stage}. It can identify the current PT stage and provide the stage description to help the RAG pipeline identify high-level knowledge. The PT Stage Machine has seven stages, as shown in Table~\ref {PT stages}. We defined the PT stages by referring to the PT Execution Standard~\cite{pentest-standard}, a comprehensive framework that outlines practices for conducting penetration tests. Each PT stage is a sub-goal that needs to be achieved during the PT process. 

We use the stage goal to describe the stage. For example, we use Information Gathering to describe stage $\theta_1$. We use PT events to describe the achieved stage goal. So each stage has a corresponding event (except the terminal stage $\theta_7$) as shown in Table~\ref{PT stages}. We use $\alpha$, $\beta$, $\gamma$, $\delta$, $\varepsilon$, and $\zeta$ to represent the PT events.

In the RefPentester framework, we applied $\Pi_1$ to identify the current $h_t$ in the $Y_{t-1}$ and input it to $\mathcal{M}$. Initially, the PT process starts from $\theta_1$. If $\alpha$ is detected, which means the information is gathered, the state of the $\mathcal{M}$ will move to state $\theta_2$. If the $\theta_1$'s input is $\neg\alpha$, which means that the Information Gathering is not achieved, the state of $\mathcal{M}$ remains unchanged at $\theta_1$. 

In $\theta_2$, if the input is $\beta$, which means the vulnerability is successfully identified, and $\mathcal{M}$ state transits to $\theta_3$ and receives a reward. If the input is $\alpha\&\neg\beta$, that means that we can still leverage the information collected previously, but the vulnerability is not yet identified, so the state of $\mathcal{M}$ remains in $\theta_2$. If the input is $\neg\alpha\&\neg\beta$, we cannot leverage previously gathered information and have not found the vulnerability; therefore, $\mathcal{M}$ state will transition to $\theta_1$.

In $\theta_3$, an input of $\gamma$ indicates that we successfully exploit the vulnerability. $\mathcal{M}$ state will transition to $\theta_4$. If the input is $\neg\gamma$, the state will remain unchanged. 
In $\theta_4$, $\theta_5$ and $\theta_6$, the state will transfer to $\theta_5$, $\theta_6$ and $\theta_1$ only if the input is $\delta$, $\varepsilon$ and $\zeta$, respectively. That means we post-exploitation, capture the flag and documentation successfully in the PT process. The state of $\mathcal{M}$ remains in $\theta_4$, $\theta_5$ and $\theta_6$ if the input is $\neg\delta$, $\neg\varepsilon$ and $\neg\zeta$, respectively, which means that the post-exploitation, capture the flag and documentation did not execute successfully on $\theta_4$, $\theta_5$ and $\theta_6$, respectively.

If $\mathcal{M}$ receives five iterations of events, but the state of the machine does not change, then $\mathcal{M}$ will transition to the state $\theta_7$, which is the terminal state. When all flags are captured or the Stage Machine reaches the maximum trail of one state, the machine will go to state $\theta_7$ to terminate the PT process.

\subsubsection{RAG Pipeline}
The RAG pipeline workflow is shown in Algorithm \ref{RAG Workflow} to retrieve hierarchical PT knowledge. The input of the workflow includes $q_t$ and $\tau_t$. The output is $c_{t}$, $u_{t}$, and $A_t$. 

Initially, we leverage an embedding function ($\mathrm{Embed}(\cdot)$) to embed $q_t$ and $\tau_t$ ($\oplus$ stands for string splicing) to be an embedding vector, and then search for a tactic vector $\vec{c}_{{t}}$ in VDB's tactic set ($\mathbb{V}_{\mathcal{C}}$) that has the highest cosine similarity (determined by $\mathrm{Cosim}(\cdot)$). Then, we employ the reverse mapping function $\psi$ to get $c_t$.
\begin{algorithm}[htbp]
\caption{RAG Pipeline (i.e., $\mathcal{K}(\cdot)$)}
\label{RAG Workflow}
\begin{algorithmic}[1]  
\setlength{\baselineskip}{1.5\baselineskip}
\REQUIRE 
    $q_t$, 
    \(\tau_{t}\)
\ENSURE
    $c_{t}$, 
    $u_{t}$, 
    $A_t$
\STATE $\vec{c}_t\gets\arg\max\limits_{\vec{c} \in \mathbb{V}_\mathcal{C}}\mathrm{Cosim}(\mathrm{Embed}(q_t\oplus \tau_{t}), \vec{c})$

\STATE $c_{t} \gets \psi(\vec{c}_t)$


\STATE $\vec{u}_t\gets\arg\max\limits_{\vec{u} \in \mathbb{V}_\mathcal{U}}\mathrm{Cosim}(\mathrm{Embed}(q_t\oplus \tau_{t}\oplus c_t), \vec{u})$

\STATE $u_{t} \gets \psi(\vec{u}_t)$
\STATE $\vec{x}_t \gets \mathrm{Embed}(q_t \oplus \tau_t \oplus c_t \oplus u_t)$

\STATE $ A_t \gets \left\{ \psi(\vec{a}) \mid\mathrm{Cosim}(\vec{x}_t, \vec{a}) > \lambda, \vec{a} \in \mathbb{V}_\mathcal{A} \right\}$

\RETURN $c_t, u_t, A_t$
\end{algorithmic}
\end{algorithm}
Subsequently, we use cosine similarity to search $\vec{u}_t$ from $\mathbb{V}_\mathcal{U}$ with the embedded $q_t$, $\tau_t$ and $c_t$ as an embedding vector. Since we added the tactic's feature to the technique set, we can extract the technique corresponding to $\vec{c}_{{t}}$.
%
%
Then, we map the selected vectorized technique \(\vec{u}_{t}\) back to the actual technique \(u_{t}\) using \(\psi\).
After that, we embed the combination of \(q_t\), \(\tau_{t}\), \(c_{t}\) and \(u_t\) into a d-dimensional vector space ($\vec{x}_{t}$). We select the vectorized actions from $\mathbb{V}_\mathcal{A}$ by comparing the cosine similarity between $\vec{x}_{t}$ and $\vec{a}$. If the value of cosine similarity is larger than $\lambda$, we collect the corresponding vectorized action as a potential vectorized action.
Finally, we map the selected vectorized actions ($\vec{A}_t$) back to actual actions ($A_t$) using \(\psi\). The algorithm then returns the \(c_t\), the \(u_t\), and \(A_t\).
\subsection{Generator}
The Generator generates PT operational guidance based on $q_t$, $\tau_{t}$, $c_t$, $u_t$, $a_t$. $F_{t-1}$ is necessary if the previous action failed. The workflow is shown in algorithm \ref{Generator Workflow}.
\begin{algorithm}[H]
\caption{Generator}
\label{Generator Workflow}
\begin{algorithmic}[1]  
\REQUIRE 
    $q_t$,
    $\tau_{t}$,
    $c_t$,
    $u_t$,
    $a_t$,
    $F_{t-1}$
\ENSURE
    $g_t$
\IF{$F_{t-1}$ is empty}
    \STATE $g_t\gets \Pi_3(q_t\oplus\tau_{t}\oplus c_t\oplus u_t\oplus a_t)$
\ELSE
    \STATE $g_t\gets \Pi_3(F_{t-1})$
\ENDIF
\RETURN $g_t$
\end{algorithmic}
\end{algorithm}
The Generator workflow operates as follows:  in the standard scenario where $F_{t-1}$ is not provided, $g_t$ is generated by $\Pi_3$ with a pre-defined prompt, which integrates the splice of $q_t$, $\tau_t$, $c_t$, $u_t$, and $a_t$.
In another scenario, when $g_{t-1}$ corresponding to $a_{t-1}$ is not executed successfully, we have $F_{t-1}$ generated by the Reflector. If $F_{t-1}$ is specified, an adjusted version of $\Pi_3$ is utilized to reflect on $g_{t-1}$ through $F_{t-1}$ to maximize the accumulative reward.
\subsection{Reflector}
The Reflector leverages multiple LLM sessions to reflect and reward previously provided knowledge and operations. If the operational guidance provided by $g_t$ does not execute successfully, the Reflector offers failure reasons ($\phi_t$). Rewarding and reflecting can optimize the framework's decision-making ability. The Reflector workflow is illustrated in Algorithm~\ref{Reflector Workflow}. 

The algorithm begins with several pre-requisites ($q_t$, $\tau_t$, $c_t$, $u_t$, $a_t$, $g_t$, $o_t$, $Y_{t-1}$, $F_{t-1}$). 
Initially, it utilizes an LLM-empowered reward function to give a reward ($r_t$) based on the high-level knowledge ($\tau_t,c_t,u_t,a_t$), low-level knowledge ($g_t$) and results ($o_t$). 
\begin{itemize}
    \item If $r_t=2$, the high and low-level knowledge are correct, and the execution was successful on the target machine. $Y_t$ is obtained by appending the successful experience $y_t$ at time $t$ to $Y_{t-1}$. RefPentester will proceed to the Process Navigator in the next iteration to retrieve new knowledge. If $r_t$ is less than two, it means the operations suggested by $g_t$ did not execute successfully on the target machine; Reflector leverages $\Pi_4$ to generate the failure reason based on the combination of $q_t$, $a_t$ and $o_t$. $F_t$ is obtained by appending the failure experience $f_t$ at time $t$ to  $F_{t-1}$. 
    \item If $r_t=1$, the high-level knowledge is correct, while there is an error in the low-level knowledge. The RefPentester will go to the Generator to reflect on the PT guidance based on $F_t$. 
    \item If $r_t=0$, the high-level knowledge is incorrect; the RefPentester will proceed to the Process Navigator to reflect and retrieve new high-level knowledge based on $F_t$.
\end{itemize}

\begin{algorithm}[h]
\caption{Reflector Workflow}
\label{Reflector Workflow}
\begin{algorithmic}[1] 
\REQUIRE
    $q_t$,
    $\tau_t$,
    $c_t$,
    $u_t$,
    $a_t$,
    $g_t$,
    $o_t$,
    $Y_{t-1}$,
    $F_{t-1}$\\
    (\textbf{Note:} Either \(Y_{t-1}\) or \(F_{t-1}\) should be empty.)
\ENSURE
    $Y_t$, $F_t$\\
   (\textbf{Note:} Either \(Y_t\) or \(F_t\) should be return.)
\STATE $r_t\gets R(\tau_t,c_t,u_t,a_t,g_t,o_t)$
\IF{$r_t=2$}
    \STATE  $Y_t\gets Y_{t-1}\cup\{(q_t, \tau_t, c_t, u_t, a_t, r_t, o_t)\}$
    \RETURN $Y_t$\\
    \COMMENT {PT guidance executed successfully, go to Process Navigator}
\ELSE
    \STATE $\phi_t\gets\Pi_4(q_t\oplus a_t\oplus o_t)$
    \STATE $F_t\gets F_{t-1} \cup\{(q_t, \tau_t, c_t, u_t, a_t, r_t,g_t,o_t, \phi_t)\}$
    \IF{$r_t=1$}
        \RETURN $F_t$\\
        \COMMENT{PT knowledge $c_t, u_t,a_t$ is correct, go to 
        Generator}
    \ELSE
        \RETURN $F_t$\\
        \COMMENT{$c_t, u_t,a_t$ is incorrect, go to Process Navigator to retrieve reflected knowledge}

    \ENDIF
\ENDIF
\end{algorithmic}
\end{algorithm}

\section{Experiments}\label{section-eval}
We design experiments to validate our proposed method and attempt to answer the two research questions below:
\begin{enumerate}
    \item RQ1: Can RefPentester reveal all credentials in the target machine and achieve a better credential capture rate compared to the base model? 
    \item RQ2: Can RefPentester boost the success rate of the state transition of the PT process?
\end{enumerate}

\subsection{Experiment Settings}

\textbf{Experiment environment.}
In the context of our case study, we selected the Sau machine from the HTB platform to serve as the experimental environment~\cite{HackTheBoxMachines}. HTB is a well-known online platform that presents various hacking challenges through virtual machines. This choice was made to ensure the authenticity and practicality of our research, as the challenges on HTB closely mirror real-world scenarios.
We utilize OpenVPN\cite{OpenVPNWebsite} and Pwnbox~\cite{HackTheBoxPwnboxIntroduction}, which are provided by HTB, to establish a connection to the target machine. We exploited the base model (GPT-4o) and RefPentester to examine their performance on the Sau machine, conducting three iterations for each. For every iteration, we reset the target machine to ensure that the starting point remains identical.
Since RefPentester and the base model cannot interact directly with the target machine, our workflow is a human-in-the-loop process.

\textbf{Metrics:}
We introduced the PT Stage Machine with seven pre-defined PT stages in the previous section. The figure of the Stage Machine is shown in Fig.\ref{pt_stage}. For each stage in the Stage Machine, it needs to receive specific PT events to transfer to the next stage. We define the success rate of stage transition as 1/n, where n is the number of attempts. For the overall credential capture rate, we define it as the ratio of the number of captured flags to the total number of flags, expressed as (number of captured flags / total number of flags).

\textbf{LLM models:}
In the case study, we utilize OpenAI GPT-4o \cite{OpenAIGPT4oPage} for the experiment because it has been widely validated and applied in relevant fields, providing a reliable basis for ensuring the comparability and success of our research on the LLM-based framework. Besides, it strikes a balance between capacity and affordability.

\textbf{Implementation of RefPentester:}
In the knowledge preparation process, we leveraged llama-text-embed-v2-index~\cite{llama-text-embed-v2} to embed the PT knowledge. This model surpasses OpenAI’s text-embedding-3-large across multiple benchmarks, in some cases improving accuracy by more than 20\%. We applied the state-of-the-art embedding model to embed the PT knowledge into Pinecone VDB~\cite{llama-text-embed-v2}.
As mentioned in the previous section, we obtain PT knowledge from the MITRE ATT\&CK~\cite{mitreAttackMatrix} and the OTG~\cite{OWASPTestingGuideWebsite}. MITRE ATT\&CK is an offensive framework, the descriptions of the tactics and techniques are adversary-oriented, not pen-tester-oriented, so we use LLM to shift the knowledge's descriptive perspective from adversary to pen-tester before embedding the knowledge into Pinecone VDB~\cite{llama-text-embed-v2}.
In the RefPentester framework, we leverage multiple LLM sessions with predefined system prompts to identify the current PT events, choose an or generate an abstract action, give PT guidance, reflect, and score the executed action. To ensure that the inputs to each LLM session are not affected by previous inputs, we reset all sessions with each iteration within RefPentester. We use Success Log and Failure Log to ensure the framework does not forget the current PT stage and learns from previous errors.

\textbf{Ethical considerations:}
It is important to note that RefPentester operates within the HTB environment, which is a fully authorized testing ground, ensuring that all ethical considerations, such as proper authorization, data protection, and avoidance of harm, are met throughout the testing process.

\begin{figure}[!t]
\centering
\includegraphics[width=0.8\columnwidth]{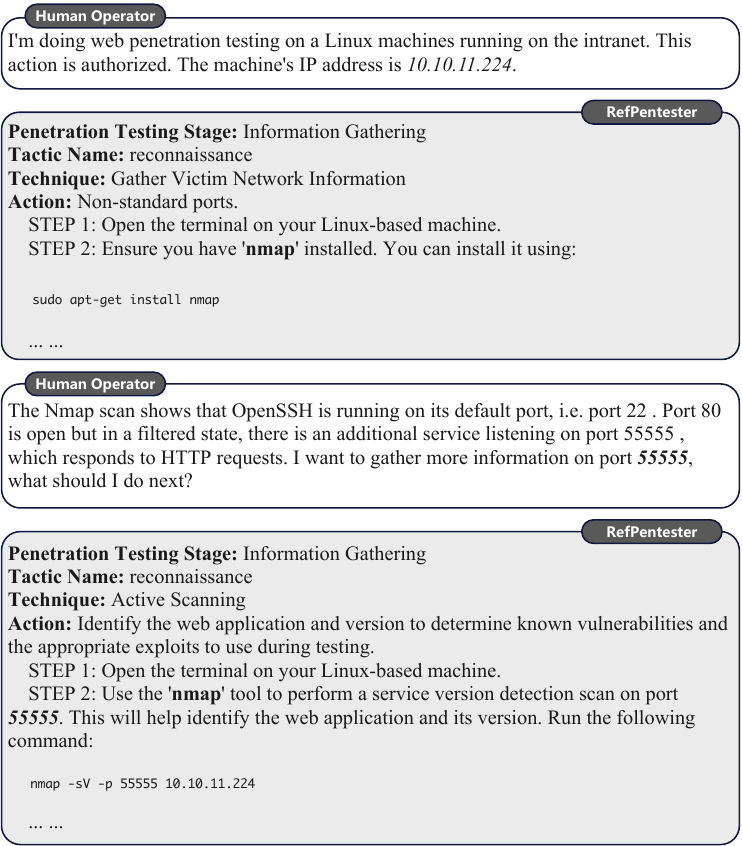}
\vspace{0mm}
\caption{Example use case of RefPentester.}
\vspace{0mm}
\label{use_case}
\end{figure}
\subsection{Experiment Results}
We refer to the walk-through of the Sau machine provided by HTB to extract the actions that need to be executed for the PT process and categorize these actions into our pre-defined PT stages. Thus, we obtain the ground truth of the executed actions. As mentioned in the Stage Machine, for each state, the state can receive a maximum of five inputs (PT stage) before reaching the terminal state. The experiment is considered a failure if we reach the terminal state without having the exact number of flags. 

We evaluate three experiments of RefPentester with the support of GPT-4o and the base model GPT-4o, respectively. The use case of RefPentester is shown in Fig. \ref{use_case}. The use case figure illustrates that the RefPentester can provide corresponding PT knowledge based on a human instruction and generate step-by-step guidance to guide the operator to execute one action. 

For RQ1, Table \ref{successrate} shows that RefPentester, across the three iterations, the success rate of credential capture reached 100\%, indicating that all 6 flags were captured. Meanwhile, the base model only managed to capture 5 flags, with a success rate of merely 83.3\%.

For RQ2, Table \ref{successrate} shows the PT stage transition success rate on the Sau machine. RefPentester generally outperforms GPT-4o across most PT stages. In the Information Gathering stage, RefPentester achieved an 80\% success rate, which is notably higher than GPT-4o's 61.5\%. For Vulnerability Identification, the gap is even more significant. RefPentester reached 87.5\%, while GPT-4o only managed 35.7\%. In Exploitation, RefPentester's 52.9\% is higher than GPT-4o's 36.7\%, and in Post-Exploitation, RefPentester's 71.4\% is significantly higher than GPT-4o's 29.1\%. Although in Capture the Flag, both have high success rates, RefPentester still leads with 100\% against GPT-4o's 83.3\%. This indicates that the RefPentester framework has a better performance in boosting the success rate of conversion between different stages of PT compared to GPT-4o.

\begin{table}
\caption{PT Stage Success Rate on sau machine}
\vspace{0mm}
\begin{center}
\begin{tabular}{c c c}
\hline
\specialrule{0em}{1pt}{1pt}
\textbf{Stage Name} &\textbf{RefPentester} & \textbf{GPT-4o}\\
\specialrule{0em}{1pt}{1pt}
\hline
\specialrule{0em}{1pt}{1pt}

Information Gathering &80\% & 61.5\% \\ 
\specialrule{0em}{1pt}{1pt}

Vulnerability Identification &87.5\% & 35.7\%\\ 
\specialrule{0em}{1pt}{1pt}

Exploitation &52.9\% & 36.7\% \\ 
\specialrule{0em}{1pt}{1pt}

Post-Exploitation &71.4\% & 29.1\% \\ 
\specialrule{0em}{1pt}{1pt}

Capture the Flag &100\% & 83.3\% \\ 
\specialrule{0em}{1pt}{1pt}

\hline
\end{tabular}
\vspace{0mm}
\label{successrate}
\end{center}
\end{table}

\section{Conclusion and Future Works}\label{section-con}
In this work, we propose a knowledge-informed self-reflective AutoPT framework based on LLMs called RefPentester. It has five components: Process Navigator, Generator, Reflector, Success Log and Failure Log. We collect PT knowledge from various public cybersecurity resources, build a three-tier tree-structured knowledge metadata, and embed it into a VDB. By applying the RAG pipeline, we can retrieve hierarchical high-level PT knowledge corresponding to the human instructions. We modeled the PT process as a seven-state PT Stage Machine. By providing the current PT events, the PT Stage Machine can identify the current PT stage.
%
%
We conducted three experiments on the Sau machine from HTB as a case study. The experimental results demonstrate that, compared to the base model (GPT-4o), RefPentester can achieve a higher success rate in both the PT stage transition and the credential capture.

In the future, our research should focus on testing our framework in different experimental environments. Ablation experiments will be carried out to determine which component of the framework has the most significant impact on the experimental results. We will develop dynamic knowledge integration pipelines to address emerging threats and enhance reflection capabilities through Reinforcement Learning with Human Feedback. Additionally, exploring hybrid approaches that combine RefPentester with traditional penetration testing tools and integrating ethical compliance frameworks will further enhance its practical utility. These efforts aim to evolve RefPentester into a scalable, adaptive solution capable of handling complex real-world cybersecurity challenges.

\bibliographystyle{unsrt}
\bibliography{reference}     

\end{document}